\documentclass{article}

\usepackage[final]{neurips_fitml_2024}

\usepackage[utf8]{inputenc} 
\usepackage[T1]{fontenc}    
\usepackage{hyperref}       
\usepackage{url}            
\usepackage{booktabs}       
\usepackage{amsfonts}       
\usepackage{nicefrac}       
\usepackage{microtype}      
\usepackage{xcolor}         
\usepackage{graphicx,subcaption}
\usepackage{natbib}
\usepackage{todonotes}
\usepackage{amsmath}

\newcommand*\samethanks[1][\value{footnote}]{\footnotemark[#1]}

\title{Balancing Cost and Effectiveness of Synthetic Data Generation Strategies for LLMs}

\author{%
  Yung-Chieh Chan\thanks{Denotes equal contribution. Work was done while Yung-Chieh was interning at Scale AI.}\hspace{1em}%
  \and \textbf{George Pu}\samethanks\hspace{1em}%
  \and \textbf{Apaar Shanker}\hspace{1em}%
  \and \textbf{Parth Suresh}\hspace{1em}%
  \and \textbf{Penn Jenks}\hspace{1em}%
  \and \textbf{John Heyer}\hspace{1em} \textbf{Sam Denton} \\
  \centerline{Scale AI} \\
  \texttt{\{yungchieh.chan, george.pu, sam.denton\}@scale.com}
}

\begin{document}

\maketitle


\begin{abstract}
As large language models (LLMs) are applied to more use cases, creating high quality, task-specific datasets for fine-tuning becomes a bottleneck for model improvement. Using high quality human data has been the most common approach to unlock model performance, but is prohibitively expensive in many scenarios. Several alternative methods have also emerged, such as generating synthetic or hybrid data, but the effectiveness of these approaches remain unclear, especially in resource-constrained scenarios and tasks that are not easily verified. To investigate this, we group various synthetic data generation strategies into three representative categories -- Answer Augmentation, Question Rephrase and New Question -- and study the performance of student LLMs trained under various constraints, namely seed instruction set size and query budget. We demonstrate that these strategies are not equally effective across settings. Notably, the optimal data generation strategy depends strongly on the ratio between the available teacher query budget and the size of the seed instruction set. When this ratio is low, generating new answers to existing questions proves most effective, but as this ratio increases, generating new questions becomes optimal. Across all tasks, we find that choice of augmentation method and other design choices matter substantially more in low to mid data regimes than in high data regimes. We provide a practical framework for selecting the appropriate augmentation method across settings, taking into account additional factors such as the scalability of each method, the importance of verifying synthetic data, and the use of different LLMs for synthetic data generation.

\end{abstract}

\section{Introduction}

Applications of large language models (LLMs) cover a wide range of tasks, from natural language understanding to code generation \citep{qin2024large, jiang2024survey}. However, applying LLMs to new tasks and domains brings challenges in sourcing high-quality, task-specific data \citep{ling2023domain}. 
To overcome this data bottleneck, various solutions have emerged, leveraging human input, hybrid methods, and synthetic data. Some examples of these approaches involve manual or automated enhancement of data quality \citep{weng2024humandata}, increasing the dataset quantity \citep{gunasekar2023textbooks, wang2024codeclm}, or extracting more informative learning signals from each data sample \citep{setlur2024rl}. For instance, \citet{dubey2024llama} enhanced smaller Llama 3.1 models in coding, math, and long-context tasks by fine-tuning on hybrid data from Llama 3.1 405b.

While each of these methods has shown promise, their relative cost-effectiveness and performance across different tasks and data constraints remain unclear, especially in resource-constrained scenarios. This lack of clarity poses a significant challenge for practitioners seeking to optimize their data generation strategies for specific tasks and available resources.

In this paper, we investigate the effectiveness of various synthetic data generation strategies for training LLMs under different constraints. We choose a knowledge distillation setting where we only have access to a set of seed instructions, a teacher LLM, and a student LLM to be fine-tuned. The objective is to leverage the limited set of seed instructions and our choice of teacher model to best improve the student model. To evaluate these strategies, we study the performance of training a student LLM under various constraints, such as seed instruction set size and query budget. The seed instruction set size represents the number of initial task-specific instructions available, while the query budget reflects the number of allowed queries to the teacher model.

Motivated by the many augmentation strategies in math domains \citep{yu2023metamath, xu2023wizardlm}, we categorize synthetic data generation methods into three main approaches -- Answer Augmentation, Question Rephrasing, and New Question -- and assess the generalizability to multiple tasks, including mathematics, coding (SQL), and general question answering. We also focus on disentangling and identifying the critical dimensions to consider when designing a data strategy for training LLMs. Our main contributions are summarized as follows:

\begin{enumerate}
    \item We propose a novel framework to evaluate synthetic data generation strategies under data constraints and demonstrate synthetic data effectiveness in new tasks beyond traditional mathematical and coding scenarios.
    \item We demonstrate that the optimal data generation strategy depends on the query budget-to-seed instruction set size ratio, where augmenting responses is most effective with limited queries, while generating new instructions becomes better as the query budget increases.
    \item We identify that model choice for New Question evolution is a key factor for student LLM performance, while showing that factors like response verification, choice of augmentation LLM for Question Rephrase, and choice of student LLM have less impact.
\end{enumerate}

\section{Related Work}

\textbf{Synthetic Data for Fine-Tuning.} Fine-tuning on synthetic and hybrid data has proven successful across a wide range of tasks \citep{liu2024best}. In the domain of mathematical reasoning, high-quality instructions are scarce, so many works leverage LLM-generated synthetic data to significantly improved the math reasoning ability of small LLMs \citep{yu2023metamath, li2024common, setlur2024rl, luo2023wizardmath}. In code generation, synthetic data from LLMs can be further verified by running test cases or the code directly, which helps close the gap between closed-source LLMs and smaller LLMs \citep{wei2024magicoder, yang2024synthesizing}. Similar approaches have been applied in instruction-following, where LLMs are effectively trained on diverse synthetic instructions with minimal to no human supervision
\citep{xu2024magpie, wang2022self, xu2023wizardlm}. However, most works focus on a single domain and do not explore how these techniques perform under varying data constraints and strategies, leaving uncertainty in adapting them to new applications. We aim to compare and extend these methods in cost constrained settings, while investigating which factors in synthetic data generation remain impactful across multiple tasks and data budgets.

\textbf{Efficient Synthetic Data Generation.} Although synthetic data is significantly cheaper than real data, its scalability encourages researchers to generate it at extremely large scales, making generation costs a substantial component of fine-tuning expert models \citep{li2024common}. Other works focus on aggressively filtering synthetic datasets for diversity and correctness with custom tricks for each domain \citep{long2024llms}. Current research on training LLMs with synthetic data emphasizes scalability and performance, but to make these methods more applicable to more tasks, we also need to disentangle and understand cost-efficiency across different scales. \citet{bansal2024smaller} explores and optimizes the choice of LLMs to sample synthetic data for overall cost reductions. Our work addresses this challenge from a new perspective by offering a general framework that guides model trainers in defining and refining their synthetic data generation strategies to maximize cost-efficiency within budget constraints.

\section{Method}
\begin{figure}
  \includegraphics[scale=0.375]{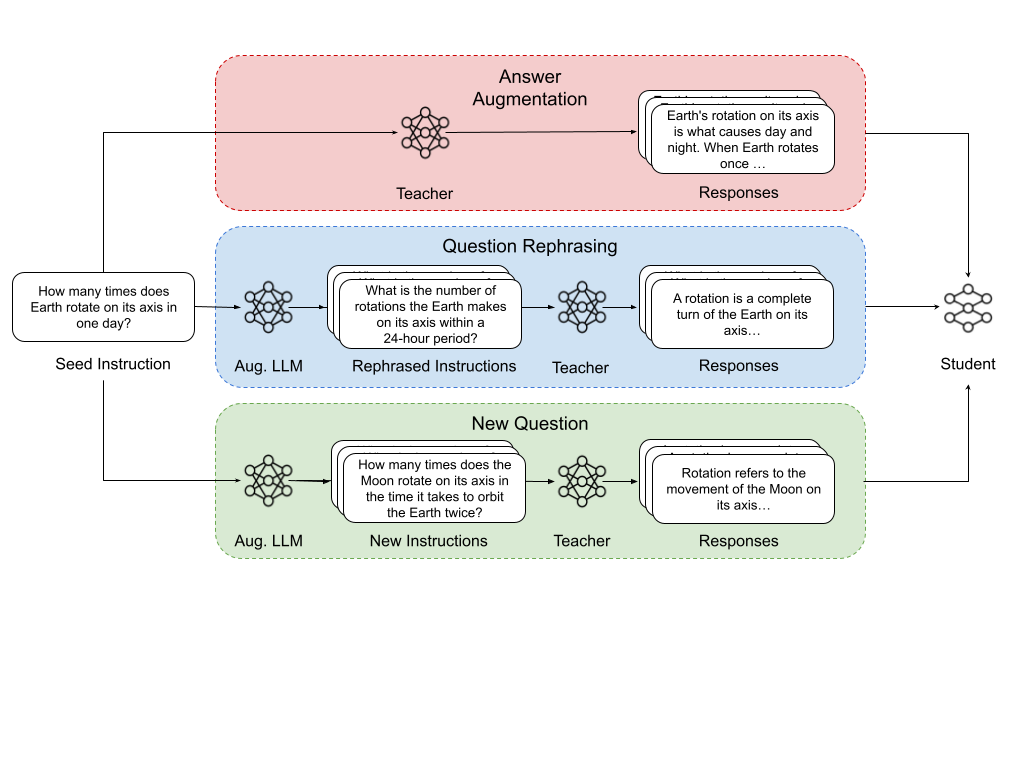}
  \vspace{-2.7cm}
  \caption{\textbf{Overview of Synthetic Data Generation Approaches}. Given a seed instruction set, we have 3 different methods to create instruction-response pairs for fine-tuning our student model. We use an example seed instruction from the ARC-C training set with synthetic instructions and responses generated with Llama 3.1 70b Instruct.}
  \label{fig:augmentation_overview}
\end{figure}
In this work, we examine techniques in synthetic data that were initially introduced in the math-reasoning domains and extend the analysis of these strategies to a more diverse collection of tasks and constraints \citep{setlur2024rl, yu2023metamath, li2024common}. We choose supervised fine-tuning (SFT) as the learning objective for the student model, which requires a dataset consisting of instruction-response pairs. We identify three main types of synthetic data generation strategies pertinent to an SFT instruction set -- Answer Augmentation, Question Rephrasing and New Question -- that are in essence, either the transformations of the seed instructions by an augmenter LLM, generation of corresponding responses by a teacher LLM or both. In this work, we do not utilize or assume the availability of ground truth responses for our instruction sets.

First, we establish relevant notations to facilitate the discussion of our data generation strategies and experiments. Let $I_{seed} = \{q_i\}_{i=1:N}$ be the set of seed or initial instructions of size $N$. Then, a synthetic data generation strategy $G$ can be understood as a two-step process: (1) augmentation of the seed instructions using an augmenter LLM referred to as $\pi_{aug}$  and (2) the generation of corresponding responses for each of these instructions using a teacher LLM, referred to as $\pi_T$. In the different choices of $G$, $\pi_{aug}$ and $\pi_T$ are utilized to obtain a synthetic training dataset $D_{synth} \in \{I_{train}, R_{train}\}$ where $I_{train}$ is the set of
instructions and $R_{train}$ is the corresponding responses towards training a student model referred to as $\pi_S$.

We present an overview of our data generation methods in Figure \ref{fig:augmentation_overview}. The following section details the construction of $D_{synth}$ using these three strategies, illustrated through the initial question: \textit{"How many times does Earth rotate on its axis in one day?"}

\subsection{Data Generation Strategies}
\label{data_generation_strategies}
\textbf{Answer Augmentation} generates a diverse set of responses to the seed instructions, varying in reasoning paths, lexical choices and semantic content. We perform Chain-of-Thought prompting with our teacher model $\pi_T$ and use temperature sampling ($\mathcal{T}$) to increase response variety \citep{wei2022chain}. Given our question about Earth's rotation, responses created by answer augmentation would generally begin with background and explanatory text: \textit{"Earth's rotation on its axis is what causes day and night..."} and end with a final answer: \textit{"The Earth rotates on its axis once in one day"}.

\vspace{-0.45cm}
\begin{align}
D_{synth} &= \{(q_i, r_i): q_i \in I_{seed}, r_i = \pi_T(q_i|\mathcal{T}=0.7), i=1:N\}
\end{align}


\textbf{Question Rephrasing} generates new instructions $I_{train}$ by reformulating the seed instructions using the augmentation model $\pi_{aug}$ before sampling the corresponding responses from the teacher model $\pi_{T}$.
Often, prompting LLMs to generate more diverse and relevant instructions is a much harder task than generating diverse solutions for a given question.
An example of a rephrased instruction would be \textit{"What is the number of rotations the Earth makes on its axis within a 24-hour period?"}, such that there is the same final answer to both this and the initial question. In practice, the rephrased questions can be paired with the responses already available in the seed instruction sets, thus providing a particularly appealing alternative from the cost-efficiency angle. However, in our study, we only assume the availability of instructions.

\vspace{-0.45cm}
\begin{align}
  D_{synth} &= \{(q_i, r_i): q_i = G_{QR}(I_{seed}|\pi_{aug}, \mathcal{T}=0.7), r_i = \pi_T(q_i|\mathcal{T}=0.7), i=1:N\}
\end{align}

\textbf{New Question Evolution} generates new instructions using $\pi_{aug}$ conditioned on examples derived from the seed instruction set but prompted to have a different final answer, and then samples $\pi_T$ for their responses. Similar to \citet{li2024common}, we adopt a self-verification process to ensure that the generated instructions are answerable and adhere to the correct format. By conditioning the $\pi_{aug}$ generation on the seed instruction set, we observe that the new instructions are semantically better aligned with the target domain. At the same time, this approach ensures more diversity in the sampled instructions compared to question rephrasing. An example of a new instruction given an initial seed instruction can be: \textit{"How many times does the Moon rotate on its axis in the time it takes to orbit the Earth two times?"}, which has a different final answer: \textit{"The Moon rotates on its axis two times."} compared to answer augmentation and question rephrase.

\vspace{-0.45cm}
\begin{align}
D_{synth} &= \{(q_i, r_i): q_i = G_{NQ}(I_{seed}|\pi_{aug}, \mathcal{T}=0.7), r_i = \pi_T(q_i|\mathcal{T}=0.7), i=1:N\}
\end{align}


\section{Experimental Setup}
\vspace{-.2cm}
\begin{table}
  \caption{Student model $\pi_S$ and teacher model $\pi_T$ accuracy on each dataset's held-out test set of 1k samples, where higher is better.}
  \label{model-performance}
  \centering
  \begin{tabular}{lccc}
    \toprule
    Model & GSM8k & Spider & ARC-C \\
    \midrule
    Llama 2 7B Chat & 15.0\%  & 23.7\%  & 47.6\%  \\
    Llama 3.1 70B Instruct & 94.6\%  & 77.3\%  & 93.9\% \\
    \bottomrule
  \end{tabular}
\end{table}

Our setup consists of a student model $\pi_S$, an augmentation model $\pi_{aug}$, a teacher model $\pi_T$ and a task-specific seed instruction set. $\pi_S$ learns from an expanded dataset with instructions generated by $\pi_{aug}$ from the seed instruction set and their corresponding responses generated by $\pi_T$. To evaluate the broad applicability of our approach, we select three diverse task types: math, coding and general question answering. We choose coding and general question answering as task types based on popular industry use cases to understand how trends in synthetic data extend beyond math-heavy domains. 

For our primary experiments, we use Llama 3.1 70b Instruct as both $\pi_T$ and $\pi_{aug}$ and Llama 2 7b Chat as $\pi_S$ \citep{dubey2024llama, touvron2023llama}. This choice of models ensures a wide performance gap between $\pi_S$ and $\pi_T$, allowing us to highlight the relative improvements to $\pi_S$ with each technique. To better understand additional cost-effective dimensions, we also run several ablations, such as the choice of $\pi_{aug}$ and $\pi_S$, in Section \ref{ablations}. Additional details on our fine-tuning process and design choices are in Appendix \ref{appendix:train_details_design_choices}.


\subsection{Datasets and Evaluations}
We select one representative dataset for each of the three tasks to evaluate the synthetic data generation strategies at scale. For math, we use GSM8k \citep{cobbe2021training}, which consists of grade school-level math questions and has 7,500 seed instructions used in training. We report the exact accuracy of the final answer for evaluation. For coding, we utilize Spider \citep{yu2018spider}, which is a text-to-SQL dataset across 200 different databases and 138 domains. We use the full 7,000 seed instructions for our experiments, and for evaluation, execution accuracy is reported as the evaluated metric. For general question answering, we use ARC-C \citep{clark2018think}, a reasoning-focused question-answering dataset consisting of grade school science questions. For each dataset, we use the respective held-out test containing about 1000 samples each to evaluate $\pi_S$. In the case of ARC-C, we evaluate our models on challenge split with exact accuracy. The initial accuracy of $\pi_S$ and $\pi_T$ are shown in Table \ref{model-performance}, while the seed instruction sets and evaluation details are described in Appendix \ref{appendix:prompts-and-eval}.

\subsection{Generating Synthetic Data under Data Constraints}
This work examines the effectiveness of the synthetic data generation strategies of interest under seed data and cost constraints for a diverse set of tasks. To simulate real-world data constraints, we create three shards of sizes 100, 1000 and "full" for each of the three source datasets. As such, we end up with nine seed instruction sets of varying sizes and task types. Let this set of seed instruction sets be denoted as $\mathcal{S} \text{ where } |\mathcal{S}|=9$. We now pair each of the generation methods ($\mathcal{G}$) with each of the seed instruction sets ($\mathcal{S}$) and perform repeated samplings at a high temperature to arrive at the final expanded training datasets of sizes $\mathcal{Q} \in \{1k,\ 2.5k,\ 5k,\ 10k,\ 25k,\ 50k,\ 100k\}$. Thus, the number of distinct training sets can then be represented as $|\mathcal{S}\times\mathcal{G}\times\mathcal{Q}| = 81$. The amount we sample $\pi_T$ and $\pi_{aug}$ towards creating a synthetic dataset leads to the notion of a query budget. Additionally, we define budget ratio (BR) as $BR = q/s$, which measures the trade-off between the query budget and the seed instruction size. A higher BR indicates more resources available for synthetic data generation relative to the initial data size. We leverage BR as a cost-sensitized metric to examine the effectiveness of different synthetic data generation strategies.


\section{Experimental Results}
For all experiments, we assess the effectiveness of each synthetic data generation strategy by comparing the accuracy of fine-tuned $\pi_S$ on the evaluation split of the dataset used in the experiment. First, we understand the effectiveness of each data generation strategy through scalability and data constraints in Section \ref{section:effectiveness_synth_data_gen_strategies}. Next, we investigate the cost-effectiveness of creating new instructions and responses in Section \ref{section:cost-effectiveness}. Finally, we cover ablation studies focused on understanding the impact of different $\pi_{aug}$, verification of responses, and a different $\pi_S$ in Subsection \ref{ablations}.


Additionally, to fit the curves in our plots and better model the scaling relationship of our data generation methods, we adopt prior work on LLM scaling laws and data-constrained scaling laws to our setting \citep{hoffmann2022training, muennighoff2024scaling}. Appendix \ref{appendix:scaling_relationship} includes additional details on how we model this scaling behavior. 

\subsection{Effectiveness of Synthetic Data Generation Strategies}
\label{section:effectiveness_synth_data_gen_strategies}
\begin{figure}
  \centering\includegraphics[scale=0.19]{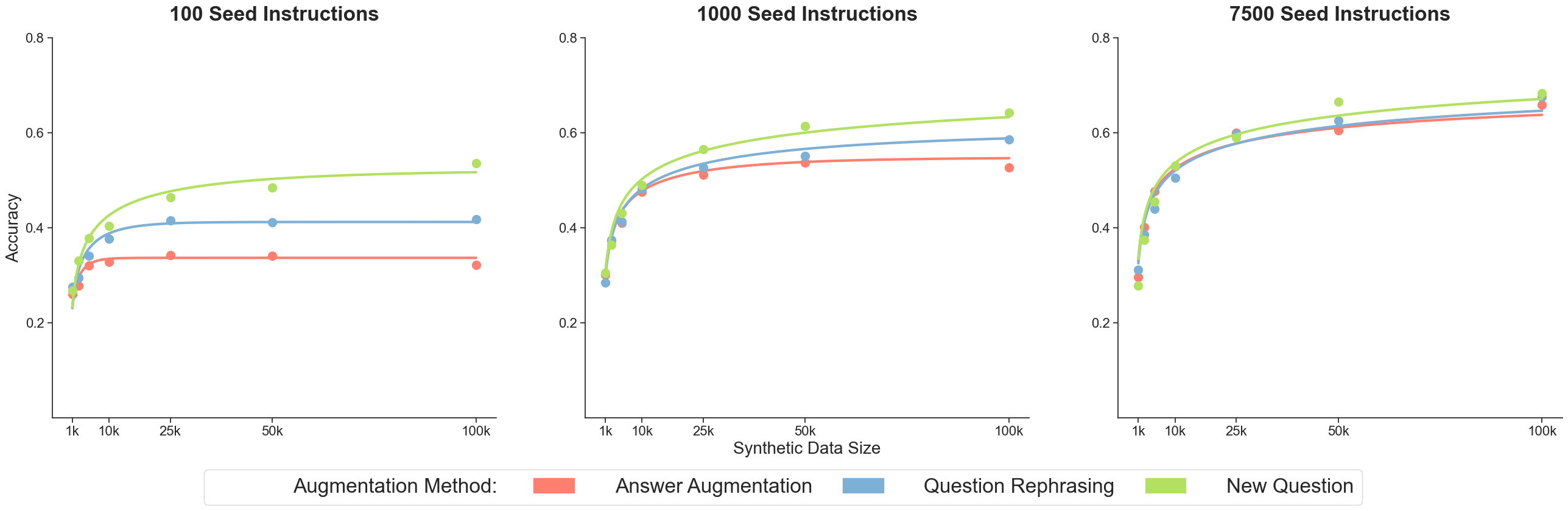}
  \centering\includegraphics[scale=0.19]{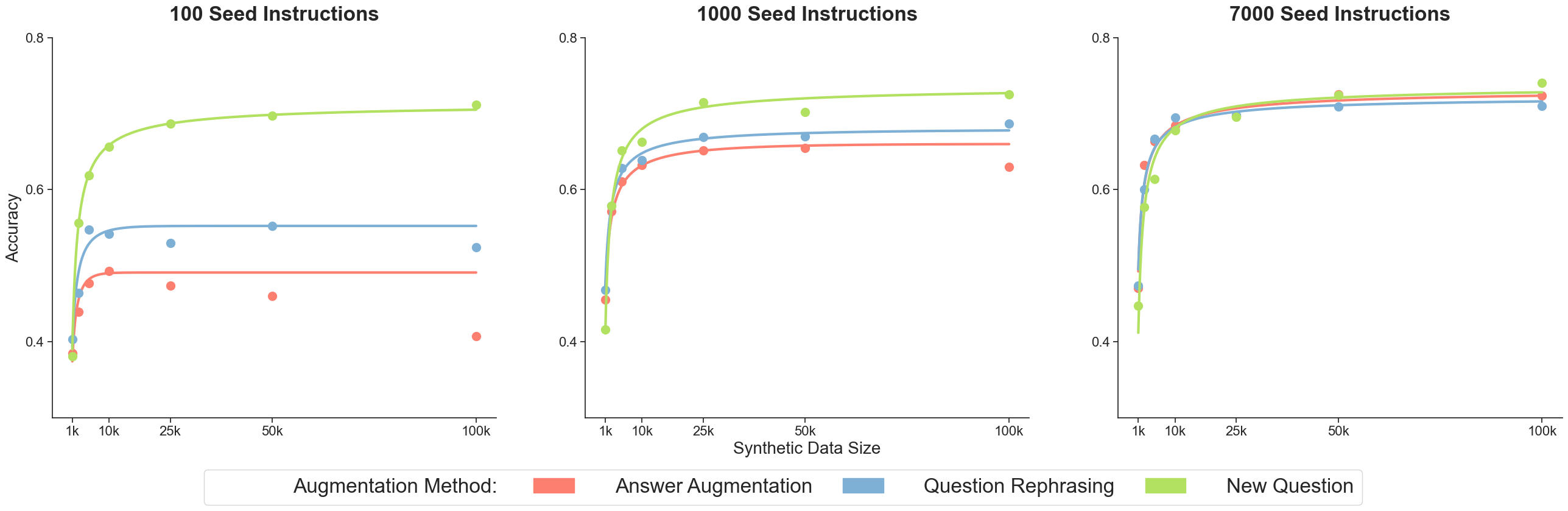}
  \centering\includegraphics[scale=0.19]{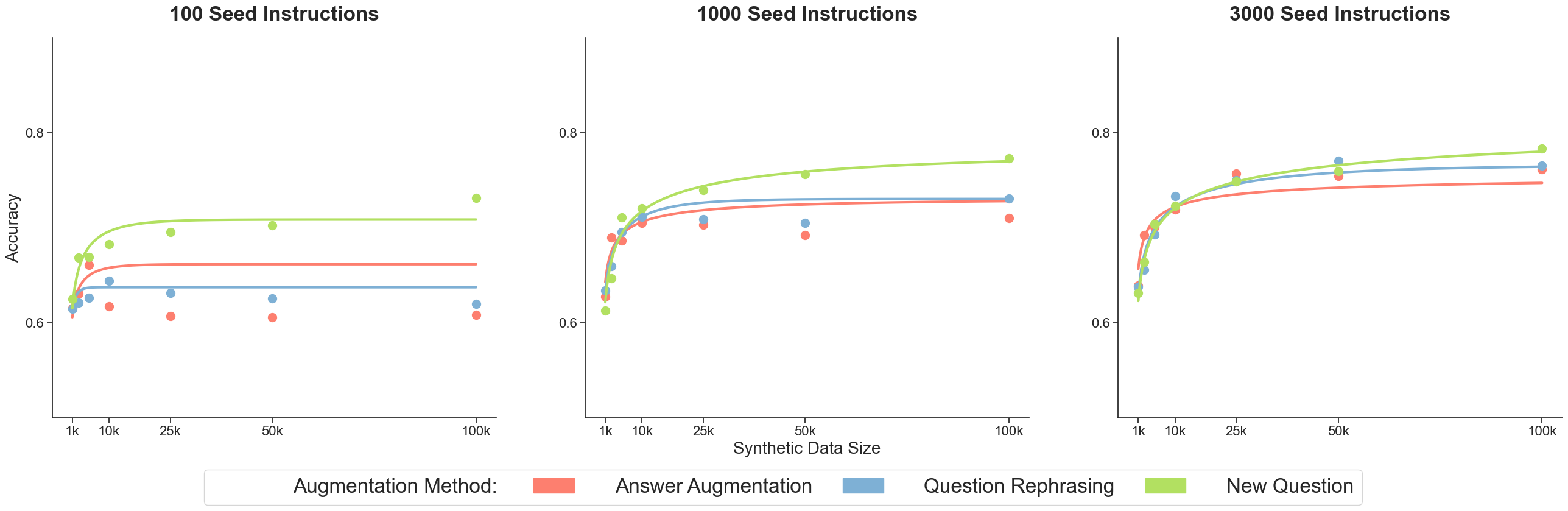}
  \caption{\textbf{Student model $\pi_S$ accuracy on GSM8k (Top), Spider (Middle) and ARC-C (Bottom)} after fine-tuning on synthetic data from our teacher model $\pi_T$ and across resource constraints.}
  \label{fig:scaling_effectiveness_plots}
\end{figure}
We study the effectiveness of data generation methods by comparing the accuracy of $\pi_S$ when fine-tuned on equal amounts of synthetic data from each method. This comparison allows us to measure the value of each added synthetic example from each generation method across different data budgets. The results for GSM8k, Spider, and ARC-C are presented in Figure \ref{fig:scaling_effectiveness_plots}. Notably, these trends generalize across data constraints, and scalability holds across our evaluated datasets.

In our GSM8k experiments with 100 seed instructions, new question evolution continues to improve accuracy as we scale the dataset beyond 50,000 examples (over 500 times the initial size) while other generation methods plateau.
However, as we increase the seed instruction set, the performance gap between augmentation methods narrows. In Spider and ARC-C, we observe the optimal augmentation method varies when our synthetic data size is between 10,000 to 50,000 samples.
The impact of data generation strategies and design choices on model performance is more pronounced in settings with fewer amounts of initial seed instructions compared to abundant data settings. This relationship becomes evident when examining our evaluation results for $\pi_S$ across various experimental configurations (Tables \ref{appendix:gsm-query-performance}, \ref{appendix:spider-query-performance}, \ref{appendix:arc-c-query-performance}).



\subsection{Cost-Effectiveness Analysis: When to Create New Instructions or Responses?}
\label{section:cost-effectiveness}
\begin{figure}
  \centering\includegraphics[scale=0.19]{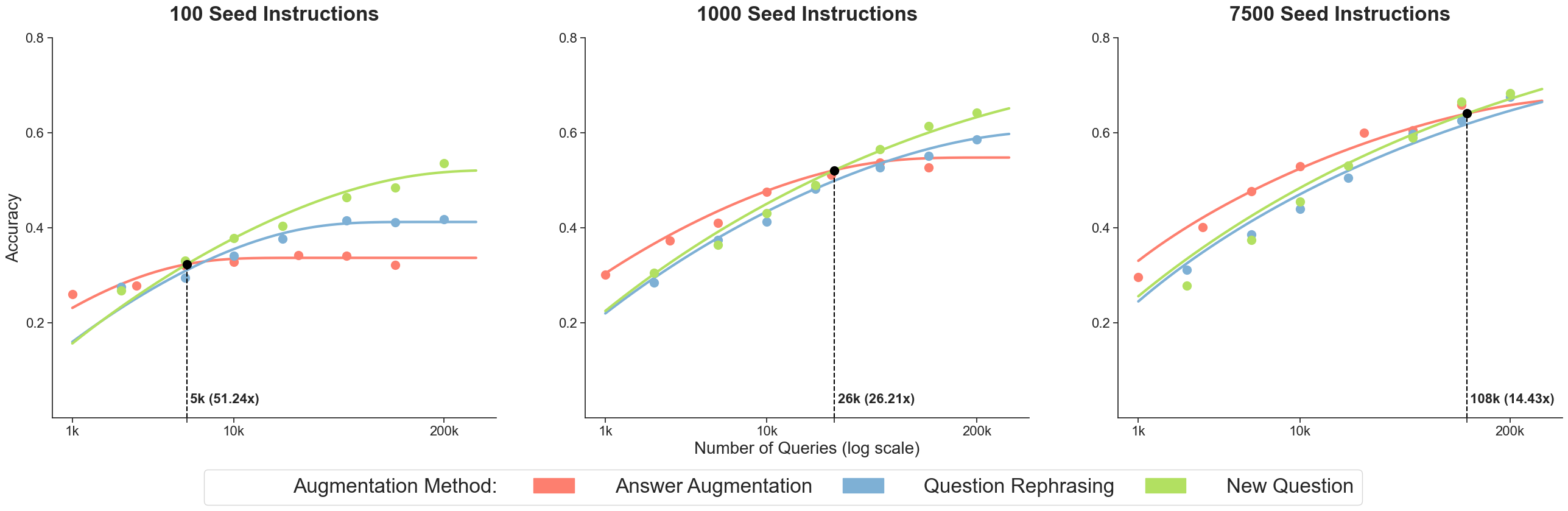}
  \centering\includegraphics[scale=0.19]{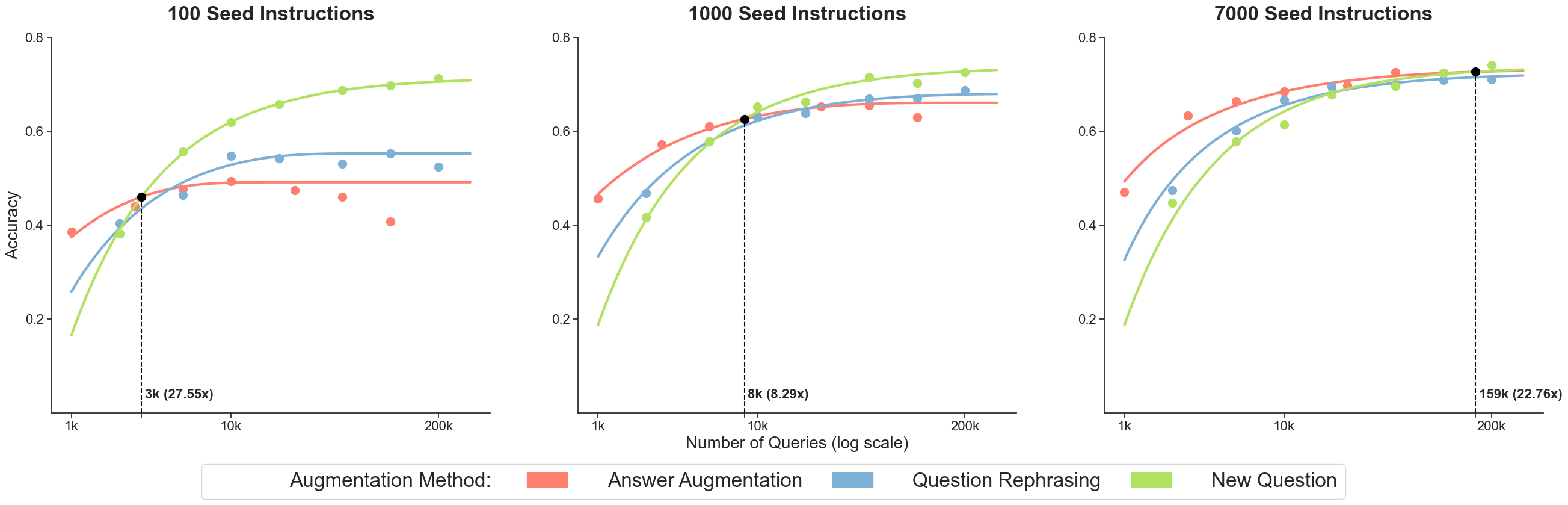}
  \centering\includegraphics[scale=0.19]{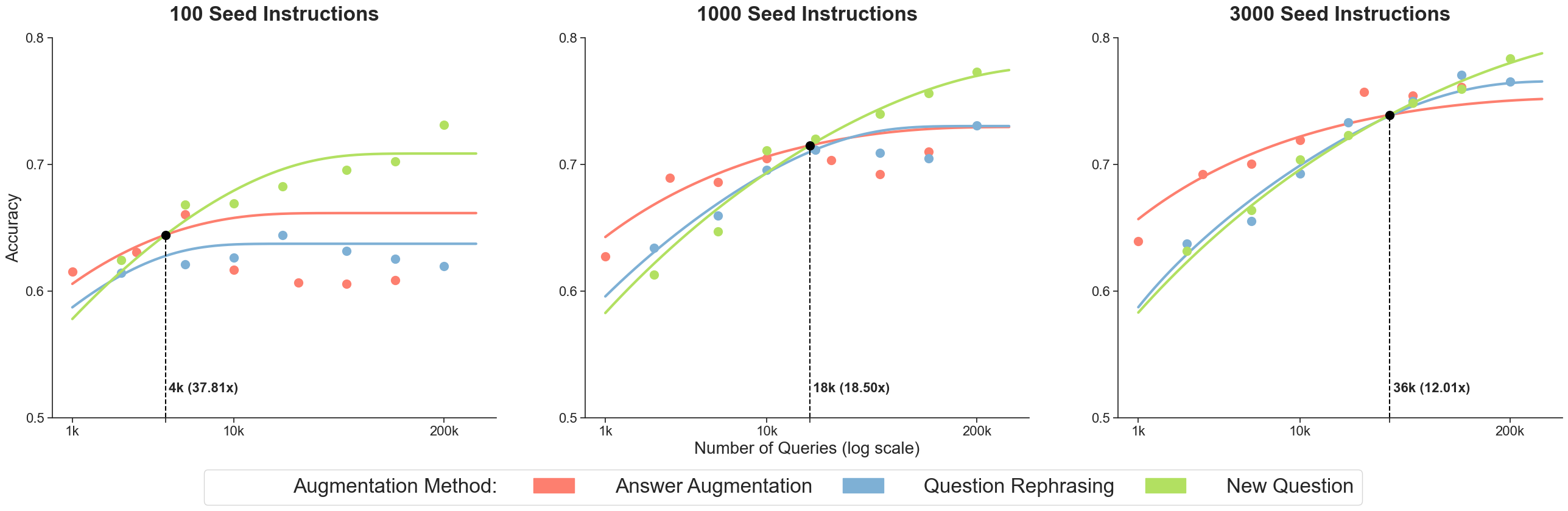}
  \caption{\textbf{Cost-Effectiveness on GSM8k (Top), Spider (Middle), and ARC-C (Bottom):} Across all three seed instruction sizes, the dashed line marks the query budget when the optimal data generation strategy changes from generating new responses to new instructions. For details on how the regression curves were fitted using our scaling relationship model, please refer to Appendix \ref{appendix:scaling_relationship}.}
  \label{fig:cost_effectiveness_overview}
\end{figure}
The optimal data generation strategy depends on many factors, such as the cost of querying $\pi_T$, final dataset size, and complexity of the task. We investigate one of these dimensions: \textit{When should we add new responses or new instructions pairs to our dataset?}

Within our cost constrained setting, answer augmentation helps us measure improvements solely from creating new responses to our initial prompts. Both question rephrase and new question represent augmentations in the prompt space. We provide empirical analysis of the optimal augmentation strategy at each budget and assume that the cost of generating a synthetic instruction-response pair using Answer Augmentation : Question Rephrase : New Question is $1:2:2$ based on the total number of queries made to $\pi_T$ and $\pi_{aug}$. In Figure \ref{fig:cost_effectiveness_overview}, we show the cost and effectiveness relationship between the three data generation methods on GSM8k, Spider and ARC-C. From our experiments with different seed sizes and tasks, all results suggest that answer augmentation is most effective option when our budget ratio (BR) is low. Thus, if we have a small query budget relative to our seed instruction size, we should spend more time creating new responses to our existing prompts. However, creating new prompts, such as through question rephrase or new question evolution, becomes the best option when the BR is high.

Generally, new question evolution outperforms question rephrase in both cost and scalability, but rephrasing questions is an easier task and may fit better in certain constrained settings.
The intersection point between these methods, which denotes the shift in optimal data strategy from new responses to new prompts, varies based on the seed instruction size. For a small seed set of 100 instructions, this point occurs at a BR between 27 and 51, corresponding to approximately 3,000 to 5,000 samples. As the seed set grows, the average BR across tasks at which this shift occurs decreases. For 1,000 seed instructions, this average BR is 17.6, whereas at our largest seed size, average BR is 16.4. These findings highlight the importance of considering both the initial dataset size and the available budget when determining the most effective data generation strategy for training LLMs.

\subsection{Ablations}
\label{ablations}
\subsubsection{Performance Trade-off with Different Augmentation Models}
\label{ablation:diff_aug_model}
We investigate whether we can reduce the cost of creating $D_{synth}$ by using a cheaper $\pi_{aug}$ when generating synthetic instructions. Given how the trends across cost-effectiveness transferred across tasks, we conduct experiments using GSM8k with 1,000 seed instructions and create $D_{synth}$ with up to 10,000 synthetic examples. We consider various choices of $\pi_{aug}$, such as Llama 2 7B, Llama 3 8B, Llama 3.1 8B, and Llama 3.1 70B, all while keeping the teacher model $\pi_T$ constant as Llama 3.1 70B and using the instruct versions.

\begin{figure}
  \centering\includegraphics[scale=0.22]{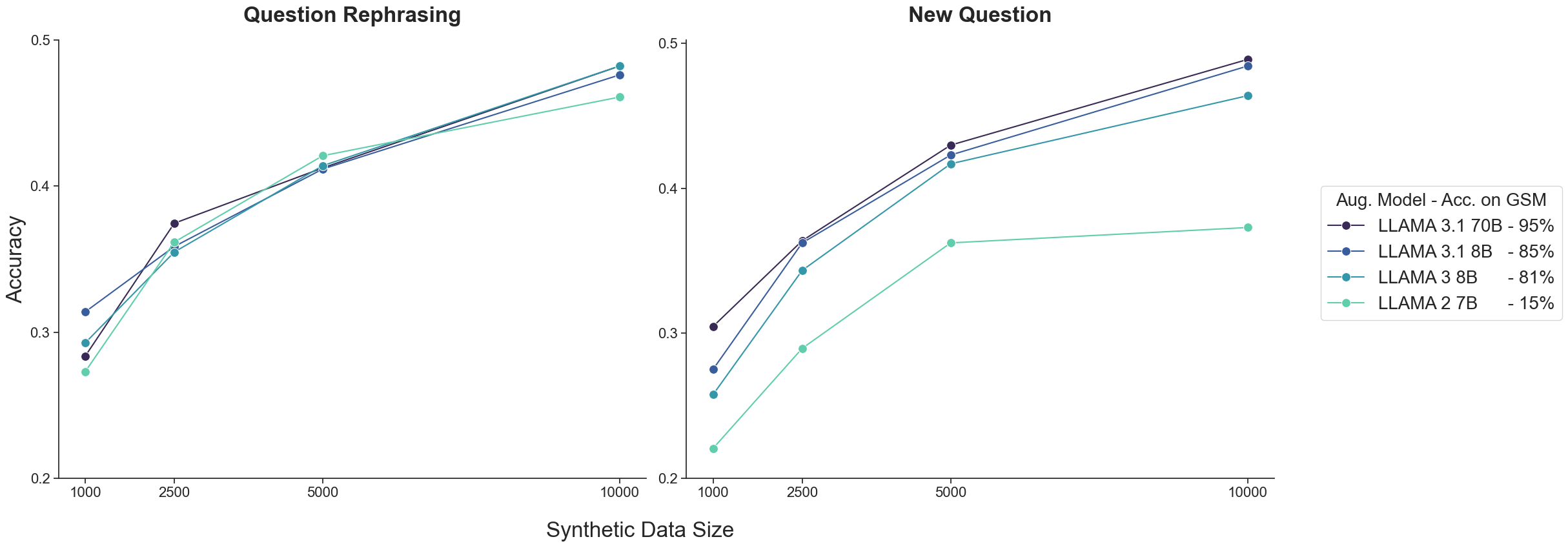}
  \caption{Performance Trade-off with Weaker Augmentation Model $\pi_{aug}$ on GSM8k}
  \label{fig:weak_aug_ablation}
\end{figure}

As illustrated in Figure \ref{fig:weak_aug_ablation}, the effectiveness of question rephrasing remains relatively robust, even when weaker augmentation models are used. However, the performance of new question evolution is closely tied to the capability of $\pi_{aug}$ on this task. At a synthetic data size of 10,000 samples, we notice a substantial decrease in performance when using Llama 2 7B, with accuracy dropping by approximately 15\% compared to other choices of $\pi_{aug}$. Since no significant degradation is seen for question rephrasing, this may be a much easier task for an LLM. We can reduce costs by using a cheaper model with this method while maintaining overall effectiveness.

\subsubsection{Effect of Verification with Ground Truth Answers}
\label{ablation:verification}
Since our teacher models $\pi_T$ produce incorrect responses for a portion of instructions in $D_{synth}$, the training process inevitably incorporates noisy and erroneous answers.
We examine whether verifying the synthetic responses from $\pi_T$ with correct answers improves the effectiveness of $D_{synth}$ when fine-tuning $\pi_S$. We focus on answer augmentation and question rephrase because the human final answers match the original or rephrased instruction, whereas new question evolution creates an unverifiable answer. Thus, we fine-tune $\pi_S$ on the Spider dataset with Llama 3.1 70B Instruct as $\pi_T$ and $\pi_{aug}$. We consider two scenarios: equal dataset and filtered dataset sizes, which addresses the possibility that the lower dataset size caused by filtering limits the effectiveness of verification.

In Figure \ref{fig:spider_verification_same_size}, we present the first scenario using 1,000 seed instructions and compare $\pi_S$ fine-tuned on an equal number of training samples at each synthetic data size after filtering for correct samples. Despite using the same number of training samples, $\pi_S$ trained on verified responses does not show significant improvement compared to the gains from scaling up the dataset size. 
In Figure \ref{fig:spider_verification}, we consider the second scenario where we query $\pi_{aug}$ and $\pi_T$ with 1,000 and 7,000 seed instructions to create a synthetic dataset up to 100,000 samples which is then filtered for good (verified) samples to fine-tune $\pi_S$. Across both data generation methods and seed instruction sizes, we similarly observe no significant improvement from verification. This can possibly be attributed to several factors. First, $\pi_S$ may still benefit from incorrect responses generated by a more capable $\pi_T$, similar to findings from \citet{yu2023metamath}, where Llama 2 still showed improvements when trained on incorrect GPT-3.5 responses. Second, verification could reduce overall diversity of $D_{synth}$. In our setting, $\pi_T$ fails to provide correct responses for 10\% of the instructions, which are filtered out and result in fewer unique instructions. 

\subsubsection{Cost-Effectiveness with a Different Student Model}
In Section \ref{section:cost-effectiveness}, we show how cost-effectiveness relationships apply across various tasks, along with the optimal data generation strategy shifting between low and high BR settings. We reconsider the choice of $\pi_S$ and assess if our cost-effectiveness results hold for a different student model. Using different data generation methods, we replicate our previous experiment with 1,000 seed instructions from GSM8k, setting our student model $\pi_S$ to Mistral 7B
\citep{jiang2023mistral}. We present our results in Figure \ref{fig:diff_student_model}, which reveals consistent cost-effectiveness patterns across methods and suggests the important factors are in the quality or diversity responses from $\pi_T$. Similarly, answer augmentation proves most effective at low budget ratios (BR), while creating new instructions becomes more beneficial as BR increases. As we see fine-tuning Mistral 7B, this transition occurs at BR of 12, aligning with our medium resource scenario (1,000 seed instructions) using Llama 2 7B, where the optimal BR fell between 8 and 26. These results reinforce the generalizability of our cost-effectiveness findings across different model architectures and sizes.




\section{Conclusion}
In this study, we provide a framework to analyze the effectiveness of various synthetic data generation strategies for training LLMs under different resource constraints and task types. Our findings reveal that the optimal strategy hinges on the ratio of the query budget to the size of the seed instruction set. Augmenting answers to existing questions proves most effective when this ratio is low, while generating new questions becomes advantageous as the ratio increases. We also find that the choice of augmentation strategy is less critical in data-rich scenarios, potentially leading to future cost reductions and efficiency improvements. Furthermore, question rephrasing is robust even with weaker augmentation models, highlighting the potential for cost reduction in specific scenarios. Finally, our observations indicate that verification of synthetic responses and the specific choice of student model have less impact. These insights should guide practitioners in selecting the most suitable data generation strategies for more efficient LLM training within their specific constraints.




\bibliography{custom}
\bibliographystyle{custom}

\clearpage
\appendix
\section{Scaling Relationship of Data Augmentation Strategies}
\label{appendix:scaling_relationship}
As we repeatedly augment and query with the same instruction set $I_{train}$, each query adds less information to $D_{synth}$, leading to a decay in per-query accuracy gain. To model this decaying effect of repeated querying, we adapt the exponential decay formulation from \citet{muennighoff2024scaling} to our settings of unique seed instructions, $S$, and repeated queries, $Q$. 

We found this exponential decay formulation fits our results significantly better than the original Chinchilla Scaling Law and other regression methods, especially on the augmentation methods that have a quick per-query effectiveness decay when querying repeatedly. This formulation model our results very closely with $R^2$ over 0.98 on all combinations of tasks and augmentation methods, and we use these $\theta$ parameters in Table \ref{synthetic-data-effectiveness} to plot our best-fit curves.

Specifically, we model the accuracy of an augmentation method given a data budget as:

$$Acc(S, Q) = E - \frac{A}{S^{\alpha}} - \frac{B}{S + S \cdot R^{*} \cdot (1 - e^{-\frac{Q/S}{R^{*}}})^{\beta})}$$

where $S$ is the size of the seed instructions, $Q$ is the number of queries we make to our teacher model, and $\theta = \{E,A,B,\alpha,\beta, R^*\}$ are learnable parameters.

On a high level, we break down efficiency into 3 factors:
\begin{enumerate}
    \item $E$: The maximum possible accuracy.
    \item $\frac{A}{S^{\alpha}}$: The part of accuracy improvable by scaling up the seed data size ($S$).
    \item $\frac{B}{S + S \cdot R^{*} \cdot (1 - e^{-\frac{Q/S}{R^{*}}})^{\beta})}$: The part of accuracy improvable by increasing the query ($Q$), with exponential decay as we repeatedly augment the same instruction.
\end{enumerate}

For each combination of task and generation method, we obtain the empirical accuracy $a_{s, q}$ of the student model by fine-tuning on $D_{synth}$ generated from $s$ seed instructions and $q$ queries. Before fitting $a_{s, q}$ to our student model results, we observed overfitting when using Answer Augmentation and Question Rephrasing at high budget ratio BR ($q/s$). Repeatedly generate from the same seed instructions cause the accuracy to decrease after $q$ increase over a certain threshold. To reflect the actual achievable accuracy under query budget $q$ and account for our monotonically increasing function $Acc$, we adjust $a_{s, q}$ to $\hat{a}_{s,q} = max_{s' \leq s, q' \leq q}\: a_{s',q'}$, which represent the highest accuracy obtainable under the given budget. We 
fit $Acc(s, q)$ by optimizing for the mean squared error:

$$min_{\theta} \sum_{\{s,q\}\in S \times Q} (\hat{a}_{s,q} - Acc_{\theta}(s, q))^2$$

with the boundary $1 \geq E \geq 0$, $A \geq 0$, and $B \geq 0$ using Limited-memory BFGS.

\begin{table}
  \caption{Effectiveness of Synthetic Data (additional figures)}
  \label{synthetic-data-effectiveness}
  \centering
  \begin{tabular}{lccccccc}
    \toprule
    Task & Generation Method & E & A & B & Alpha & Beta & R* \\
    \midrule
    GSM8k     & Answer Aug.   & 1.000 & 0.401 & 1.856 & 0.144 & 0.176 & 27.012 \\
            & Question Rephrase & 1.000 & 0.391 & 1.915 & 0.175 & 0.170 & 79.123 \\
            & New Question   & 1.000 & 0.511 & 2.005 & 0.272 & 1.699 & 385.956 \\
    \midrule
    Spider  & Answer Aug.   & 1.000 & 0.544 & 16.089 & 0.083 & 0.616 & 21.681 \\
            & Question Rephrase & 0.771 & 0.771 & 29.544 & 0.266 & 0.708 & 44.206 \\
            & New Question   & 0.738 & 9.350 & 60.586 & 1.316 & 0.756 & 320.083 \\
    \midrule
    ARC-C     & Answer Aug.   & 1.000 & 0.361 & 1.519  & 0.055 & 0.378 & 53.265 \\
            & Question Rephrase & 1.000 & 0.024 & 0.878  & 0.076 & 0.131 & 11.276 \\
            & New Question   & 0.996 & 0.109 & 1.020  & 1.632 & 1.887 & 80.799 \\
    \bottomrule
  \end{tabular}
\end{table}

\section{Related Work: Efficient LLM Training}
As model sizes and data requirements grow exponentially, optimizing the training process for Large Language Models (LLMs) has become increasingly critical. Researchers have investigated pre-training scaling laws and data mixtures, guiding model trainers toward more efficient pre-training strategies \citep{hoffmann2022training, muennighoff2024scaling, kaplan2020scaling, sachdeva2024train}. Noticing the increasing capabilities of open-source LLMs and the growing demand for task-specific LLMs, several works have explored efficient methods during post-training. From a computational efficiency perspective, studies on parameter-efficient fine-tuning demonstrate techniques to reduce the compute requirements for fine-tuning \citep{pu2023empirical, han2024parameter}. Likewise, for data efficiency, previous works have successfully reduced data requirements in the fine-tuning process by upsampling for quality \citep{zhou2024lima, ivison2022data}. These works inspire our investigation of cost and efficiency when fine-tuning with LLM-generated synthetic data.


\section{Ablation: Effect of Verification Plots}

\begin{figure}
  \centering\includegraphics[scale=0.24]{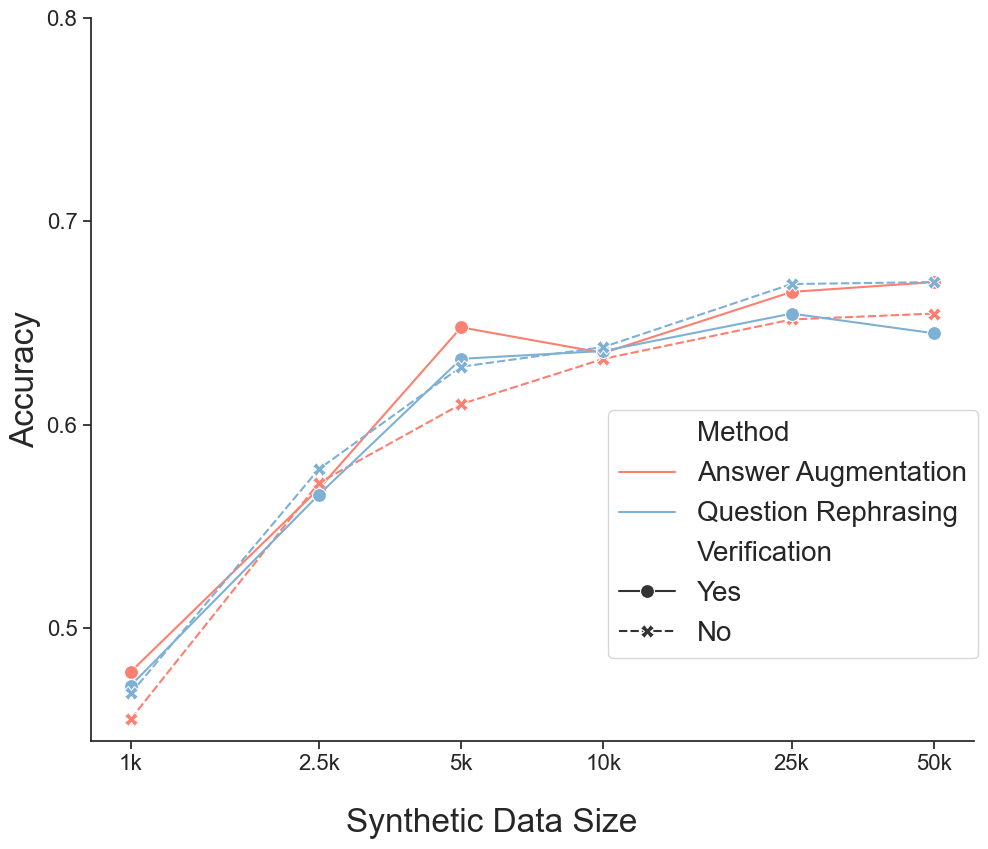}
  \caption{Ablations measuring the effect of verification with 1,000 seed instructions from Spider. We ensure the synthetic data size is the amount after filtering.}
  \label{fig:spider_verification_same_size}
\end{figure}

\begin{figure}
  \includegraphics[scale=0.23]{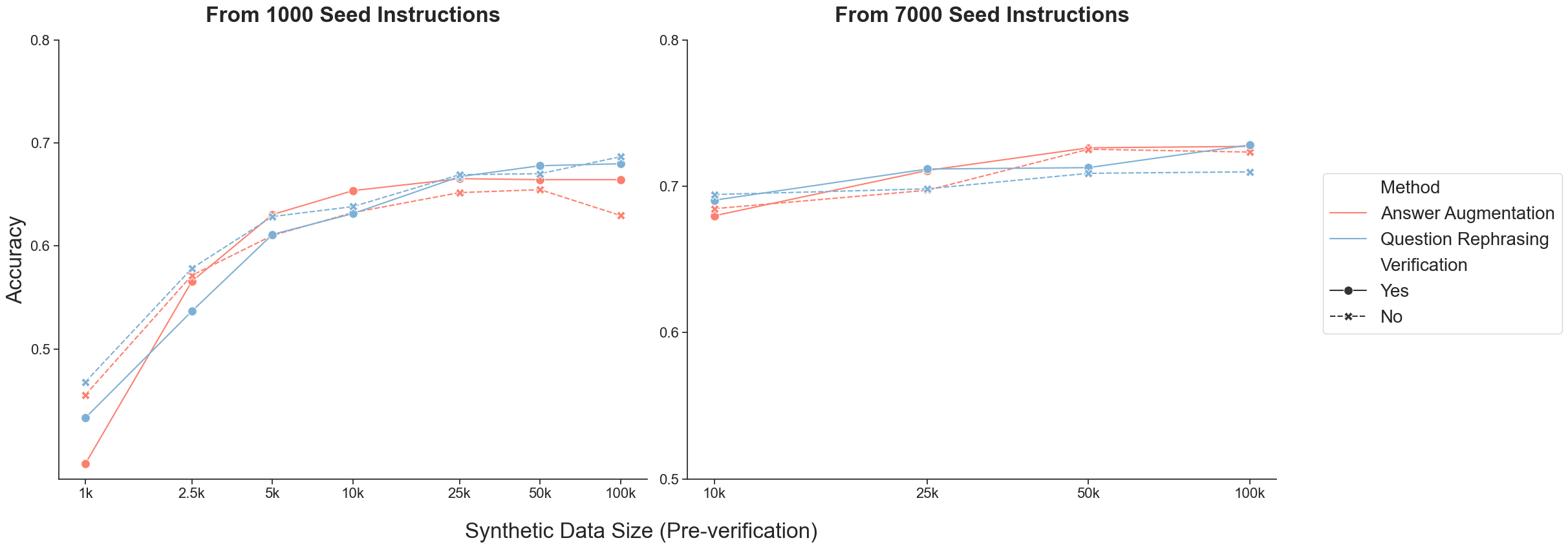}
  \caption{Effect of Verification on Spider across medium and high seed instruction sizes. The synthetic data size amount is before filtering for correctness.}
  \label{fig:spider_verification}
\end{figure}

In this section, we include both figures on our verification experiment results. In Figure \ref{fig:spider_verification_same_size}, we fine-tune $\pi_S$ on an equal number of training samples, after filtering for correctness, to each synthetic data size and evaluate the performance scaling up from 1,000 samples.

We present Figure \ref{fig:spider_verification} as another setting where we filtered out all incorrect responses from the synthetic data generated by Answer Augmentation and Question Rephrasing using 1,000 and 7,000 seed instructions, and fine-tuned $\pi_S$ on the filtered datasets. 

\section{Ablation: Different Student Model Plots}
\begin{figure}
  \centering\includegraphics[scale=0.27]{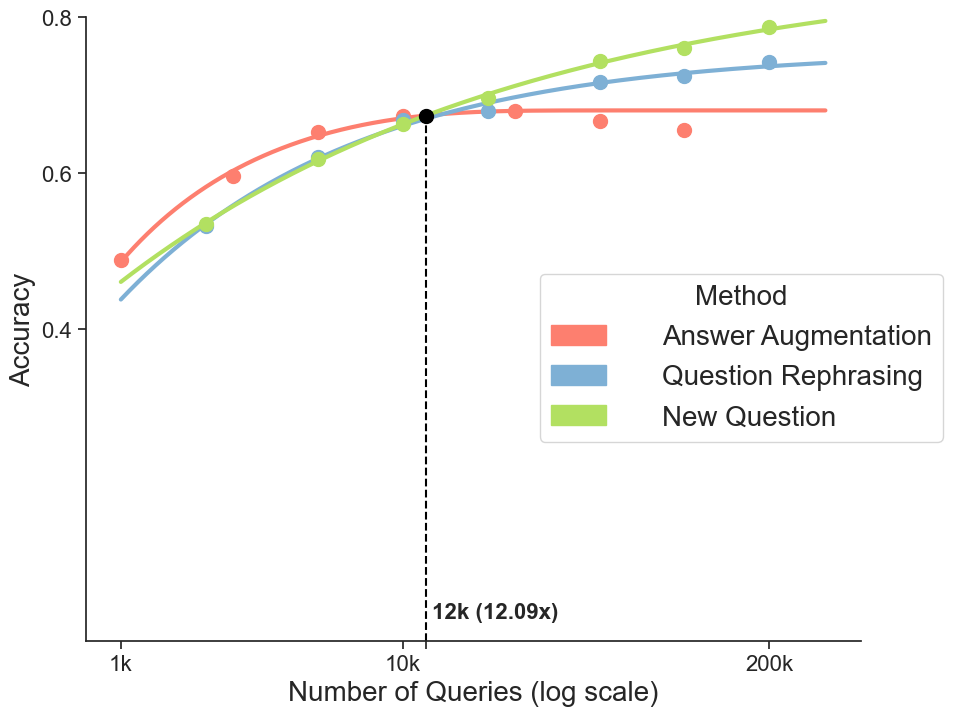}
  \caption{Transferability of Cost-Effectiveness on 1k seed instructions from GSM8k with Mistral 7B as $\pi_S$}
  \label{fig:diff_student_model}
\end{figure}

We present Figure \ref{fig:diff_student_model} to showcase the transferability of our cost-effective analysis with Llama 2 7b to a different student model.

\section{Accuracy Scores Across Tasks}
\label{appendix:total_accuracy_tables}
We include the accuracy of the student model $\pi_S$, fine-tuned on the $D_{synth}$ synthetic training dataset across query budgets $(\{1k,\ 2.5k,\ 5k,\ 10k,\ 25k,\ 50k,\ 100k\})$, and generation methods $(\{Answer\ Augmentation,\ Question\ Rephrase,\ New\ Question\})$. Each task, and respective dataset, has different seed sizes $S$, but we ensure that low-resource has $S = 100$, medium-resource has $S = 1,000$ and high-resource scenario has the constraint: $1k < S \leq 10k$.

We present the results on GSM8k in Table \ref{appendix:gsm-query-performance}, which includes the performance of $\pi_S$ on $D_{synth}$ across seed sizes $\{100, 1k, 7.5k\}$. Similarly, for Spider, Table \ref{appendix:spider-query-performance} includes the performance of across seed sizes $\{100, 1k, 7k\}$. For ARC-C, Table \ref{appendix:arc-c-query-performance} includes the performance across seed sizes $\{100, 1k, 3k\}$.

\begin{table}
  \caption{Performance Across Different Query Sizes and Seeds for GSM8k}
  \label{appendix:gsm-query-performance}
  \centering
  \begin{tabular}{llccccccc}
    \toprule
    Generation Method & Seed \textbackslash \ Query & 1k & 2.5k & 5k & 10k & 25k & 50k & 100k \\
    \midrule
    & 100  & 25.9 & 27.7  & 32.0  & 32.8  & 34.2  & 34.1  & 32.1  \\
    Answer Aug. & 1000 & 30.1 & 37.3  & 41.0  & 47.5  & 51.1  & 53.7  & 52.7  \\
    & 7500 & 29.6 & 40.1  & 47.7  & 52.8  & 59.9  & 60.5  & 65.8  \\
    \midrule
    & 100  & 27.5 & 29.4  & 34.0  & 37.7  & 41.5  & 41.2  & 41.7  \\
    Question Rephrase & 1000 & 28.4 & 37.5  & 41.2  & 48.2  & 52.7  & 55.0  & 58.5  \\
    & 7500 & 31.1 & 38.6  & 43.9  & 50.5  & 59.7  & 62.5  & 67.6  \\
    \midrule
    & 100  & 26.8 & 33.0  & 37.8  & 40.3  & 46.3  & 48.3  & 53.5  \\
    New Question & 1000 & 30.5 & 36.39 & 42.98 & 48.9  & 56.5  & 61.3  & 64.1  \\
    & 7500 & 27.8 & 37.4  & 45.4  & 53.1  & 58.9  & 66.5  & 68.2  \\
    \bottomrule
  \end{tabular}
\end{table}

\begin{table}
  \caption{Performance Across Different Query Sizes and Seeds for Spider}
  \label{appendix:spider-query-performance}
  \centering
  \begin{tabular}{llccccccc}
    \toprule
    Generation Method & Seed \textbackslash Query & 1k & 2.5k & 5k & 10k & 25k & 50k & 100k \\
    \midrule
    & 100 & 38.49 & 43.9  & 47.68 & 49.32 & 47.39 & 46.03 & 40.72 \\
    Answer Aug. & 1000 & 45.56 & 57.16 & 61.03 & 63.25 & 65.18 & 65.47 & 63.0  \\
    & 1000 & 47.00 & 63.25 & 66.34 & 68.47 & 69.73 & 72.53 & 72.34 \\
    \midrule
    & 100 & 40.33 & 46.42 & 54.73 & 54.16 & 53.00 & 55.22 & 52.42 \\
    Question Rephrase & 1000 & 46.80 & 57.83 & 62.86 & 63.83 & 66.92 & 67.02 & 68.67 \\
    & 7000 & 47.39 & 60.06 & 66.64 & 69.44 & 69.83 & 70.89 & 70.89 \\
    \midrule
    & 100 & 38.10 & 55.61 & 61.90 & 65.67 & 68.66 & 69.73 & 71.18 \\
    New Question & 1000 & 41.59 & 57.83 & 65.18 & 66.24 & 71.47 & 70.21 & 72.53 \\
    & 7000 & 44.68 & 57.76 & 64.22 & 67.79 & 69.54 & 72.43 & 74.08 \\
    \bottomrule
  \end{tabular}
\end{table}

\begin{table}
  \caption{Performance Across Different Query Sizes and Seeds for ARC-C}
  \label{appendix:arc-c-query-performance}
  \centering
  \begin{tabular}{llccccccc}
    \toprule
    Generation Method & Seed \textbackslash Query & 1k & 2.5k & 5k & 10k & 25k & 50k & 100k \\
    \midrule
    & 100   & 61.51 & 63.05 & 66.04 & 61.68 & 60.66 & 60.58 & 60.83 \\
    Answer Aug. & 1000  & 65.87 & 67.49 & 71.16 & 72.10 & 72.35 & 73.38 & 73.29 \\
    & 3000  & 63.90 & 69.19 & 70.05 & 71.92 & 75.68 & 75.42 & 76.10 \\
    \midrule
    & 100   & 61.43 & 62.11 & 62.62 & 64.41 & 63.13 & 62.54 & 61.94 \\
    Question Rephrase & 1000  & 63.39 & 65.95 & 69.53 & 71.16 & 70.90 & 70.47 & 73.03 \\
    & 3000  & 63.73 & 65.52 & 69.28 & 73.29 & 75.00 & 77.04 & 76.53 \\
    \midrule
    & 100   & 62.45 & 66.80 & 66.89 & 68.25 & 69.53 & 70.22 & 73.12 \\
    New Question & 1000  & 61.26 & 64.67 & 71.07 & 72.01 & 73.97 & 75.59 & 77.30 \\
    & 3000  & 63.13 & 66.38 & 70.39 & 72.26 & 74.82 & 75.93 & 78.32 \\
    \bottomrule
  \end{tabular}
\end{table}

\section{Training Details and Design Choices}
\label{appendix:train_details_design_choices}
We used the instruction-tuned versions of all language models in this paper because they are easier to prompt for our specific tasks and produce better-formatted outputs, making them easier to parse and process. Fine-tuning was carried out over 3 epochs with a peak learning rate of 4e-5, except for the Mistral 7B model, which used a learning rate of 1e-5. A cosine decay schedule was applied, with 3\% of the total steps allocated for warm-up. The batch size was set to 128, and the maximum sequence length was 1,536 tokens. These settings were determined through a hyperparameter search on learning rate and batch size, conducted on the GSM8k training data, with the assumption that the optimal configuration would generalize to other tasks and synthetic data. All training experiments were performed on two NVIDIA H100 GPUs.

For synthetic data generation, we followed settings from math literature, using greedy decoding with a temperature of 0.7. Our exploratory experiments confirmed this as a stable choice. Lower temperatures reduced the diversity of the generated data, making fine-tuning on repeated samples less effective, while higher temperatures decreased quality without improving effectiveness. For the choice of teacher model, we compared the effectiveness of fine-tuning using responses generated by GPT-4o and Llama 3.1 70B on the GSM8k dataset and observed similar outcomes. Given that Llama 3.1 70B is cheaper and faster, we selected it as the teacher model and generated all synthetic data using vLLM \citep{kwon2023efficient} on four NVIDIA H100 GPUs.

\section{Prompts and Evaluation Details}
\label{appendix:prompts-and-eval}
In this section, we present the prompts used to generate synthetic instruction and responses across generation methods and tasks. We adapt prompts from prior work in math reasoning into all of our representative tasks -- math, coding (SQL), and general question answering. We perform additional prompt engineering to ensure the generated data resembles the original instructions. We validate this with small experiments to compare the effectiveness of training on synthetic data against real data. Similar to the experiment described in the "Comparison of Synthetic SFT Data versus Real Data" section in \citet{li2024common}, we first generate synthetic data with size equal to the original training data. Then, we train Llama 2 7B model on the synthetic data and ensure the synthetic data maintains a level of effectiveness comparable to the real data.

For our initial evaluation results in Table \ref{model-performance}, we evaluate Llama 2 with few-shot prompts pulled from the training dataset while Llama 3 is evaluated with a zero-shot COT prompt used in Answer Augmentation. We observed Llama 2 being unable to generate reasonable CoT responses without few-shot examples provided, whereas, Llama 3 scored fairly high with zero-shot examples and had fairly similar scores. Additionally, we generate synthetic responses in a zero-shot manner, so we opted to capture accuracy under the same settings.

\subsection{Answer Augmentation}
Table \ref{tab:prompts-ans-aug} contains all prompts we use for answer augmentation.

\begin{table}[htbp]
    \centering
    \begin{tabular}{|p{0.95\textwidth}|}
        \hline
        \multicolumn{1}{|c|}{\textbf{GSM8K (Math)}} \\
        \hline
        Please act as a professional math teacher. Your goal is to accurately solve a math word problem.\\[0.5em]
        To achieve the goal, you have two jobs.\\
        \# Write a detailed solution to a given question.\\
        \# Write the final answer to this question.\\[0.5em]
        You have two principles to do this.\\
        \# Ensure the solution is step-by-step.\\
        \# Ensure the final answer is just a number (float or integer).\\[0.5em]
        Given question: \{\textbf{question}\}\\
        Your output should be in the following format:\\
        SOLUTION: <your detailed solution to the given question>\\
        FINAL ANSWER: <your final answer to the question with only an integer or float number>\\
        \hline
        \multicolumn{1}{|c|}{\textbf{Spider (Coding)}} \\
        \hline
        You are a database engineer. Given database table descriptions and a question, your goal is to write a SQL query to answer the question.\\[0.5em]
        To achieve the goal, you have two jobs.\\
        \# Write a detailed solution with step-by-step reasoning to break down the problem.\\
        \# Write the final SQL query.\\[0.5em]
        You have two principles to do this.\\
        \# Ensure the solution is step-by-step.\\
        \# Ensure the final answer is only the executable SQL query.\\[0.5em]
        Given question: \{\textbf{question}\}\\
        Your output should be in the following format:\\
        SOLUTION: <your detailed solution to the given question>\\
        FINAL ANSWER: <your final executable SQL query>\\
        \hline
        \multicolumn{1}{|c|}{\textbf{ARC-C (General QA)}} \\
        \hline
        Your goal is to solve a reasoning problem. You will be given a question, followed by multiple answer choices. Only one of the choices is correct.\\[0.5em]
        To achieve the goal, you have two jobs.\\
        \# Write a detailed solution with step-by-step logical reasoning to the given question.\\
        \# Write the final chosen answer.\\[0.5em]
        Given question: \{\textbf{question}\}\\
        Your output should be in the following format:\\
        SOLUTION: <your solution to the given question>\\
        FINAL ANSWER: <your final answer choice label>\\
        \hline
    \end{tabular}
    \caption{Answer Augmentation prompts across tasks (math, coding, general qa).}
    \label{tab:prompts-ans-aug}
\end{table}

\subsection{Question Rephrasing}
Table \ref{tab:prompts-question-rephrase} contains all prompts we use for question rephrasing.

\begin{table}[htbp]
    \centering
    \begin{tabular}{|p{0.95\textwidth}|}
        \hline
        \multicolumn{1}{|c|}{\textbf{GSM8K (Math)}} \\
        \hline
        Please act as a professional math teacher. Your goal is to create high quality math word problems to help students learn math.\\[0.5em]
        You will be given a math question. Please rephrase the given question to create a new question.\\
        \# Ensure the rephrased question has the same meaning as the given question and can be answered with the same solution as the given question.
        \# Please DO NOT include solution in your question.\\[0.5em]
        Given question: \{\textbf{question}\}\\
        Your output should be in the following format:\\
        REPHRASED QUESTION: <your rephrased question>\\
        \hline
        \multicolumn{1}{|c|}{\textbf{Spider (Coding)}} \\
        \hline
        You are a professional computer science teacher. Your goal is to create high quality SQL problems to help students learn.\\[0.5em]
        You will be given a problem that contains database table descriptions and a question. Please rephrase the given problem to create a new problem.\\
        \# Ensure the rephrased problem has the same meaning as the given problem and can be answered with the same SQL query as the given problem.\\
        \# Ensure the table description is included in the rephrased problem as the same format as the given problem.\\[0.5em]
        Given problem: \{\textbf{question}\}\\
        Your output should be in the following format:\\
        REPHRASED PROBLEM: <your rephrased problem>\\
        \hline
        \multicolumn{1}{|c|}{\textbf{ARC-C (General QA)}} \\
        \hline
        Your goal is to create high quality reasoning problems to help AI learn about our world.\\[0.5em]
        You will be given a multiple choice question. Please rephrase the Given Question to create a new question.\\
        \# Ensure the rephrased question has the same meaning as the Given Question and can be answered with the same solution as the Given Question.\\
        \# Ensure the answer choices are included in the rephrased question as the same format as the given question.\\[0.5em]
        Given question: \{\textbf{question}\}\\
        Your output should be in the following format:\\
        REPHRASED QUESTION: <your rephrased question and answer choices>\\
        \hline
    \end{tabular}
    \caption{Answer Augmentation prompts across tasks (math, coding, general qa).}
    \label{tab:prompts-question-rephrase}
\end{table}

\subsection{New Question}
Table \ref{tab:prompts-new-question} and Table \ref{tab:prompts-new-question-cont} contains all prompts we use for new question evolution.

\begin{table}[htbp]
    \centering
    \begin{tabular}{|p{\textwidth}|}
        \hline
        \multicolumn{1}{|c|}{\textbf{GSM8K (Math)}} \\
        \hline
        Please act as a professional math teacher. Your goal is to create high quality math word problems to help students learn math. You will be given a math question. Please create a new question based on the given question and the following instructions.\\[0.5em]
        To achieve the goal, you have three jobs.\\
        \# Please generate a similar but new question according to the given question.\\
        \# Check the question by solving it step-by-step to find out if it adheres to all principles.\\
        \# Modify the created question according to your checking comment to ensure it is of high quality.\\[0.5em]
        You have five principles to do this.\\
        \# Ensure the new question only asks for one thing, be reasonable, be based on the given question, and can be answered with only a number (float or integer). For example, DO NOT ask, ‘what is the amount of A, B and C?’.\\
        \# Ensure the new question is in line with common sense of life. For example, the amount someone has or pays must be a positive number, and the number of people must be an integer.\\
        \# Ensure your student can answer the new question without the given question. If you want to use some numbers, conditions or background in the given question, please restate them to ensure no information is omitted in your new question.\\
        \# Please DO NOT include the solution in your question.\\
        \# If the created question already follows these principles upon your verification, just keep it without any modification.\\[0.5em]
        Given question: \{\textbf{question}\}\\
        Your output should be in the following format:\\
        CREATED QUESTION: <your created question>\\
        VERIFICATION AND MODIFICATION: <solve the question step-by-step and modify it to follow all principles>\\
        FINAL CREATED QUESTION: <your final created question>\\
        \hline
        \multicolumn{1}{|c|}{\textbf{Spider (Coding)}} \\
        \hline
        You are a professional computer science teacher. Your goal is to create high quality SQL problems to help students learn. You will be given a problem that contains database table descriptions and a question. Please create a new problem similar to the given problem with the following instructions.\\[0.5em]
        To achieve the goal, you have three jobs.\\
        \# Please generate a similar but new problem according to the given problem.\\
        \# Check if the problem adheres to all principles.\\
        \# Modify the created problem according to your checking comment to ensure it is of high quality.\\[0.5em]
        You have three principles to do this.\\
        \# Ensure the created problem has the same difficulty as the given problem.\\
        \# Ensure the created problem has a table description section in the same format as the given problem.\\
        \# Ensure the new problem can be answered without the given problem. If it uses the same table as the given problem, make sure to include them in the table description of the new problem.\\
        \# If the created problem already follows these principles upon your verification, just keep it without any modification.\\[0.5em]
        Given problem: \{\textbf{question}\}\\
        Your output should be in the following format:\\
        CREATED PROBLEM: <your created problem>\\
        VERIFICATION AND MODIFICATION: <verify the problem adheres to the principles, if not modify it to follow all principles>\\
        FINAL CREATED PROBLEM: <your final created problem>\\
        \hline
    \end{tabular}
    \caption{New Question prompts across math and coding tasks.}
    \label{tab:prompts-new-question}
\end{table}

\begin{table}[htbp]
    \centering
    \begin{tabular}{|p{\textwidth}|}
        \hline
        \multicolumn{1}{|c|}{\textbf{ARC-C (General QA)}} \\
        \hline
        Your goal is to create high quality reasoning problems to help AI learn about our world. You will be given a multiple choice question. Please create a new question based on the given question and the following instructions.\\[0.5em]
        To achieve the goal, you have three jobs.\\
        \# Please generate a similar but new question according to the given question.\\
        \# Check the question by solving it step-by-step to find out if it adheres to all principles.\\
        \# Modify the created question according to your checking comment to ensure it is of high quality.\\[0.5em]
        You have four principles to do this.\\
        \# Ensure there's only one correct answer.\\
        \# Ensure the new question can be answered without the given question.\\
        \# Ensure the answer choices are included in the created question in the same format as the given question.\\
        \# If the created question already follows these principles upon your verification, just keep it without any modification.\\
        \# Please DO NOT include the solution in your question.\\[0.5em]
        Given question: \{\textbf{question}\}\\
        Your output should be in the following format:\\
        CREATED QUESTION: <your created question>\\
        VERIFICATION AND MODIFICATION: <solve the question step-by-step and modify it to follow all principles>\\
        FINAL CREATED QUESTION: <your final created question and answer choices>\\
        \hline
    \end{tabular}
    \caption{New Question prompts across general question answering tasks.}
    \label{tab:prompts-new-question-cont}
\end{table}

\end{document}